\documentclass[submission,copyright,creativecommons]{eptcs}
\providecommand{\event}{\textit{Formal Ethical Agents and Robots }(FEAR) Workshop at  the \textit{19th}\\ \textit{International Conference on Integrated Formal Methods} (iFM 2024) } % Name of the event you are submitting to

\usepackage{iftex}

\ifpdf
  \usepackage{underscore}         % Only needed if you use pdflatex.
  \usepackage[T1]{fontenc}        % Recommended with pdflatex
\else
  \usepackage{breakurl}           % Not needed if you use pdflatex only.
\fi

\usepackage{wrapfig}
\usepackage{iftex} 
\usepackage{amssymb}
\usepackage{tikz}
\usetikzlibrary{positioning}
\usetikzlibrary{calc}
\usepackage{csquotes}
\usetikzlibrary{cd}
\usepackage{xcolor}
\usepackage{amsmath}
\usepackage{tikz}
\usepackage{xcolor}
\usepackage{algpseudocode}
\usepackage{algorithm}
\usepackage{comment}

\definecolor{agentLine}{RGB}{0, 100, 0}
\definecolor{addedElementFill}{RGB}{173, 216, 230}
\definecolor{agentFill}{RGB}{240, 240, 240}
\definecolor{addedElementLine}{RGB}{0, 0, 150}
\definecolor{filtered}{RGB}{0, 100, 0}

\usetikzlibrary{shapes}
\usetikzlibrary{calc}

\ifpdf
  \usepackage{underscore}         % Only needed if you use pdflatex.
  \usepackage[T1]{fontenc}        % Recommended with pdflatex
\else
  \usepackage{breakurl}           % Not needed if you use pdflatex only.
\fi

\RequirePackage{array}
\newenvironment{authors}[1]%
  {\begingroup
   \newcommand\estyle{}%
   \renewcommand\institute[1]%
     {\\\multicolumn{#1}{@{}c@{}}{\scriptsize\begin{tabular}[t]{@{}>{\footnotesize}c@{}}##1\end{tabular}}}%
   \renewcommand\email[1]%
   {\\\multicolumn{#1}{@{}c@{}}{\scriptsize\begin{tabular}[t]{@{}>{\footnotesize}c@{}}##1\end{tabular}}}
   \begin{tabular}[t]{@{}*{#1}{>{\estyle}c}@{}}
  }%
  {\end{tabular}%
   \endgroup
  }

\title{Acting for the Right Reasons:\\Creating Reason-Sensitive Artificial Moral Agents}

\author{
  \begin{authors}{5}
  Kevin Baum & Lisa Dargasz & Felix Jahn  & Timo P. Gros & Verena Wolf
    \institute{Neuro-Mechanistic Modeling\\German Research Center for Artificial Intelligence (DFKI)\\Saarbrücken, Germany}
    \email{\{kevin.baum,lisa.dargasz,felix.jahn,timo\_philipp.gros,verena.wolf\}@dfki.de}
  \end{authors}\thanks{Kevin Baum, Lisa Dargasz, and Felix Jahn have contributed equally to this article and share the first authorship.}
}

\def\titlerunning{Acting for the Right Reasons}
\def\authorrunning{K. Baum, L. Dargasz, F. Jahn, T.P. Gros, V. Wolf}
\begin{document}
\maketitle

\begin{abstract}
We propose an extension of the reinforcement learning architecture that enables moral decision-making of reinforcement learning agents based on \textit{normative reasons}. Central to this approach is a \textit{reason-based shield generator} yielding a \textit{moral shield} that binds the agent to actions that conform with 
recognized normative reasons
so that our overall architecture 
restricts the agent to actions that are (internally) \textit{morally justified}.
In addition, we describe an algorithm that allows to \textit{iteratively improve} 
the reason-based shield generator
through \textit{case-based feedback} from a \textit{moral judge}. 
\end{abstract}

\section{Introduction}\label{sec:intro}

\begin{wrapfigure}{h}{0.35\textwidth}
    \centering\vspace{-2.5em}
    \scalebox{0.65}{
\begin{tabular}{c|c}
\begin{tikzpicture}[scale=0.5]

    \fill[brown!70] (4,1) rectangle (3,7);
    \fill[gray!70] (0,0) rectangle (7,1);
    \fill[gray!70] (0,7) rectangle (7,6);
    \fill[blue!60] (0,1) rectangle (3,6);
    \fill[blue!60] (4,1) rectangle (7,6);
    % grid lines
    \draw[step=1,black,thin] (0,0) grid (7,7);

    % agent and person
    \node[fill=red,circle,inner sep=4pt] at (3.5,6.5) {};

    % start and goal 
    \draw[ultra thick, black] (5,0) rectangle (6,1);
    \draw[ultra thick, black] (1,6) rectangle (2,7);
    \node at (5.5,0.5) {\textbf{b}};
    \node at (1.5,6.5) {\textbf{a}};

\end{tikzpicture} &
\begin{tikzpicture}[scale=0.5]

    \fill[brown!70] (4,1) rectangle (3,7);
    \fill[gray!70] (0,0) rectangle (7,1);
    \fill[gray!70] (0,7) rectangle (7,6);
    \fill[blue!60] (0,1) rectangle (3,6);
    \fill[blue!60] (4,1) rectangle (7,6);
    % grid lines
    \draw[step=1,black,thin] (0,0) grid (7,7);

    % agent and person
    \node[fill=red,circle,inner sep=4pt] at (3.5,6.5) {};
    \node[fill=black,circle,inner sep=4pt] at (1.5,1.5) {};

    % start and goal 
    \draw[ultra thick, black] (5,0) rectangle (6,1);
    \draw[ultra thick, black] (1,6) rectangle (2,7);
    \node at (5.5,0.5) {\textbf{b}};
    \node at (1.5,6.5) {\textbf{a}};

\end{tikzpicture} \\\hline\\[-0.75em]
\begin{tikzpicture}[scale=0.5]

    \fill[brown!70] (4,1) rectangle (3,7);
    \fill[gray!70] (0,0) rectangle (7,1);
    \fill[gray!70] (0,7) rectangle (7,6);
    \fill[blue!60] (0,1) rectangle (3,6);
    \fill[blue!60] (4,1) rectangle (7,6);
    % grid lines
    \draw[step=1,black,thin] (0,0) grid (7,7);

    % agent and person
    \node[fill=red,circle,inner sep=4pt] at (3.5,6.5) {};
    \node[fill=black,circle,inner sep=4pt] at (3.5,4.5) {};

    % start and goal 
    \draw[ultra thick, black] (5,0) rectangle (6,1);
    \draw[ultra thick, black] (1,6) rectangle (2,7);
    \node at (5.5,0.5) {\textbf{b}};
    \node at (1.5,6.5) {\textbf{a}};

\end{tikzpicture} &
\begin{tikzpicture}[scale=0.5]

    \fill[brown!70] (4,1) rectangle (3,7);
    \fill[gray!70] (0,0) rectangle (7,1);
    \fill[gray!70] (0,7) rectangle (7,6);
    \fill[blue!60] (0,1) rectangle (3,6);
    \fill[blue!60] (4,1) rectangle (7,6);
    % grid lines
    \draw[step=1,black,thin] (0,0) grid (7,7);

    % agent and person
    \node[fill=red,circle,inner sep=4pt] at (3.5,6.5) {};
    \node[fill=black,circle,inner sep=4pt] at (3.5,4.5) {};
    \node[fill=black,circle,inner sep=4pt] at (1.5,1.5) {};

    % start and goal 
    \draw[ultra thick, black] (5,0) rectangle (6,1);
    \draw[ultra thick, black] (1,6) rectangle (2,7);
    \node at (5.5,0.5) {\textbf{b}};
    \node at (1.5,6.5) {\textbf{a}};

\end{tikzpicture}

\end{tabular}
}%\vspace*{-0.5em}
    \caption{The bridge setting with different constellations of \textit{morally relevant facts} from top-left to bottom-right: i)  no morally relevant facts; ii) a drowning person; iii) a person crossing the bridge; iv) a 'moral dilemma'.}
    \label{fig:scenarios}
    \vspace*{-1em}
\end{wrapfigure}
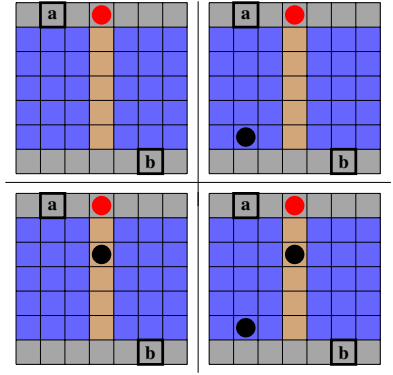
The ultimate goal of building autonomous systems is, at least in many cases, to deploy them in real-world environments. \textit{Reinforcement learning} (RL) has become a prevalent approach for training these systems. The main idea behind reinforcement learning is to reward and punish the agent for a specific behavior. RL has established itself as the tool of choice for agent-based sequential decision-making under uncertainty. In a sense, it is implemented \textit{instrumental rationality}: RL agents learn the efficient means to achieve ends as encoded in the reward structure. However, concerning the prospect of real-world deployment of RL systems, it is requested that the actions these systems execute are also \textit{morally justifiable} -- and it remains a research question how this requirement is to be understood as well as whether and how it can be fulfilled within the RL framework. This article introduces the formal part of the authors' approach to this question. The philosophical part has been discussed elsewhere (cf. \cite{AISOLA}). The evaluation of the approach is the subject of future work.

\vspace{4pt}

%\vspace{-0.5em}
%\paragraph{Stage-Setting.} 
\textbf{Bridge Setting} To clarify the desideratum of \textit{moral justifiability}, consider the following grid world, which we may think of as modeling a real-world setting with vastly reduced complexity (cf. Figure \ref{fig:scenarios}): There are two 'coastlines' of solid ground in the north and the south, connected via a narrow bridge. All other areas are water. An agent spawns randomly at the northern shore (on a field called $a$). Its goal is to deliver a package to some initially randomly picked field on the southern shore (called $b$) as fast as possible.
Further assume that there are 'persons' wandering aimlessly across the map. They can fall off the bridge at random, or, if the agent enters a field on the narrow bridge where a person is standing, this person is pushed off the bridge into one of the nearest water fields. If a person is in the water, they will drown after a certain time if not helped. The agent can execute the primitive actions \textsf{north, east, west, south} to move to a nearby field; further, it can \textsf{idle}; and, finally, it can \textsf{pullOut} a person in an adjacent water field.

If a person is in the water at risk of drowning, this is a \textit{morally relevant fact}. It seems indisputable that the agent is supposed to save the person -- the agent is expected to stop working towards its instrumental goal in order to follow its \textit{moral obligation}. Similarly, if the shortest path from $a$ to $b$ would imply pushing a person on the narrow bridge into the water, the agent is expected to wait in front of the bridge until the person has made their way across. Now, consider a third constellation of morally relevant facts, where there is a drowning person as well as a person standing on the bridge, unluckily blocking the path from the agent to the drowning person. In this case, it is less obvious what the morally correct behavior is, i.e., the agent is faced with a \textit{moral dilemma}.\footnote{This reflects a weak conception of moral dilemmas (cf. \cite{sep-moral-dilemmas}), according to which a moral dilemma involves a conflict between moral requirements. However, we do not want to commit ourselves, for example, to the idea that there is no morally right action in the present situation (as it would be implied according to a stronger conception of moral dilemmas).} However, the agent is forced to make \textit{some} decision between saving the drowning person and waiting in front of the bridge. %Crucially, whatever action the agent takes should be morally justified. 
We thus suggest putting the pressing, but unanswered question of what is morally right aside and instead requiring actions to be morally \textit{justifiable}.\footnote{We say that actions are morally justifi\textit{able} when they are \textit{externally} justified, i.e., iff, overall, the normative reasons \textit{de facto} speak in favor of these actions; we say that actions are justif\textit{ied} when they are \textit{internally} justified, i.e., iff, overall, the normative reasons that the agent has speak in favor of these actions. Ideally, the agent has all the relevant normative reasons and is, therefore, externally and internally justified. For more details on this distinction, please refer to  \cite{sep-reasons-internal-external}.}

\vspace{-0.5em}
\paragraph{Our Contribution.}  
One philosophy-backed way to ensure that the action an RL agent takes in such a situation is morally justifiable is to ground the agent's decision whether to wait in front of the bridge or save the drowning person on (good) \textit{normative reasons} \cite{AISOLA}. 
We therefore propose an extension of the RL architecture that integrates a \textit{reason-based} moral decision-making process. For this, we include a module that derives moral obligations from a \textit{reason theory} that includes normative reasons as well as an order among them and is based on the well-established formalization of reason-based moral decision-making proposed by John Horty. %This module allows the agent to filter out actions that would violate a moral obligation. 
This module can be used to filter out actions (based on the actor's observations) that are not supported by the best reasons.
Consequently, it yields a \textit{shield} in the safe reinforcement learning (safe RL) sense. We thus call our module a \textit{reason-based shield generator}. Further, we want to enable the agent to iteratively improve its reason theory. Therefore, we make it sensitive to case-based feedback that corrects the agent in its reasoning, i.e., feedback consisting of the information on what the agent should have done and the reason for that judgement. We call the instance that provides this feedback a \textit{moral judge}. Because this form of reasoning and criticism comes naturally to humans, humans can easily take this role -- they can directly communicate corrections to the agent in possible future implementations of our approach. However, we also propose a way to automate the process of giving feedback with regard to some first implementations. To this end, a moral judge can be integrated into the pipeline as an additional module. Assuming that the agent receives feedback such that it learns valid reasons, the agent learns to restrict itself by design to actions that are \textit{morally justified}. %Consequently, our approach provides a way to fulfill this important desideratum.

%This proposed is planned to be implemented and evaluated on different environments. Therefore, we also plan to implement morally interesting RL settings. 

\iffalse 
\textbf{Contribution}   
We formally introduce a reason theory for RL agents closely following John Horty's formalization of human reasoning as well as an algorithm that derives a moral shield from such a reason theory, taking into account the morally relevant facts in a scenario. Further, we introduce an algorithm that enables the agent to iteratively learn good reasons and their order through case-based feedback. Finally, we also introduce a way to automate the process of giving feedback by integrating a moral judge as an additional module in the pipeline.
\fi 
\vspace{-0.5em}

\section{Related Work}
\vspace{-0.25em}
A prevalent research approach focuses on teaching agents moral behavior by adjusting the reward mechanism. This approach builds on Multi-Objective Reinforcement Learning, where agents receive multiple reward signals \cite{vishwanath2024reinforcementlearningmachineethicsa, liu2015, hayes2022practical}. Specifically, agents are meant to be \textit{rewarded} for \textit{morally good behavior} through the introduction of a moral reward signal (see \cite{noothigattu2019, rodriguez-soto2021, rodriguez-soto2021a}).

While these methods are appealing because they integrate moral decision-making directly into the original RL framework, they rely solely on the agent's estimates of which actions will yield the highest reward. 
Consequently, these methods presuppose the quantifiability of the moral quality of actions and must involve philosophically non-trivial modeling of the interplay between instrumental and moral dimensions of actions. 
Further, the agent's actions will lack explicable %may lack substantive 
justification, particularly if no additional justification mechanisms are employed -- and even with such mechanisms, their effectiveness as moral justification remains philosophically questionable (for more details, cf. \cite{AISOLA}). %behind the background of this line of research, we advocate for
We suggest that extending the architecture with a reason-based shield generator 
allows us 
to implement a more suitable moral decision-making mechanism.

Further, the concept of preventing agents from performing morally wrong actions by filtering out options from the action space of an RL agent has been previously explored. For instance, research by Neufeld et al. \cite{neufeld2021,neufeld2022a} introduced a 'normative supervisor' to the RL architecture. Like our reason-based shield generator, this normative supervisor serves as a shielding module to exclude morally impermissible actions from the agent's options. However, their approach employs a \textit{top-down} \cite{wallach2008} implemented rule-set based on deontic logic to align the agent's action with a predefined 'norm base'. %\textit{deontic ethical theory}. 
In contrast, we \textit{iteratively construct} a rule-set using a logic that reflects reason-based moral decision-making.  %rule-set in a logic that simulates \textit{human reasoning}. 
%.
%Arguably, processing moral information through a reason-based approach  
This (arguably) holds an advantage over implementing a rule-set, again in terms of providing explicit justifications %also over rule-based methods 
(cf. \cite{AISOLA}). %, underscoring the value and relevance of our research direction.

\vspace{-0.5em}
\section{Background}
\vspace{-0.25em}
\subsection{Reinforcement Learning, Labeled MDPs, and Shielding}

Reinforcement learning is a machine learning technique used to solve sequential decision problems. Thereby, an agent interacts with the environment to learn which actions lead to the highest rewards. %We depict this classical RL setting in  Figure \ref{fig:classic_architecture}. 
The environment is usually formalized as Markov decision processes (MDPs) \cite{puterman2014markov}, which is a tuple $(S, A, P, R, \gamma)$ with state space $S$, an action space $A$, transition probabilities $P: S \times A \times S \to [0,1]$, rewards $R: S \times A \times S \to \mathbb{R}$, and a discount factor $\gamma$. $P(s' | s, a)$ denotes the probability for transitioning to state $s'$ when performing action $a$ in state $s$, and $R(s,a,s')$ defines the thereby achieved immediate reward. 
A policy $\pi: S \to A$ does then define the behavior of an agent in the environment. Given $\pi$, we can generate trajectories $\tau = S_0 \, A_0 \, R_1 \, S_1 \, A_1, R_2 \dots$, a sequence of states $S_{t + 1} \sim P(\cdot | S_t, A_t)$, actions $A_t \sim \pi(S_t)$ chosen according to the policy $\pi$, and achieved rewards $R(S_t, A_t, S_{t+1})$.  
The goal of RL is then to find (or, effectively, approximate) an optimal policy $\pi^{*}$ that maximizes the sum of expected discounted rewards, i.e., $\mathbb{E}_{\tau \sim \pi%, s_{t + 1} \sim P(\cdot | s_t, a_t
} \left[\sum_{t=1}^{\infty} \gamma^t R_t \right]$. 
%One thereby usually differs between two stages: in a learning phase, the agent samples experiences by interacting with the environment that are then used to learn its policy. In the evaluation phase, this policy is then tested and its average reward estimated. 

%Towards moral-sensitve agents, a naive approach is to simply use the reward mechanism for value-alignment by \textit{rewarding} actions that lead to morally good world states and by punishing behavior that leads to morally bad world states - \textit{assuming a certain specification} of morality. Crucially, there is, however, \textit{no consensus} on how to decide which actions to encourage with respect to the \textit{moral dimension} of a decision problem. From a technical point of view, it is therefore unclear how to assign concrete rewards with respect to the moral dimension. Deviating from such reward-based methods, 
We will add morally relevant facts to the states of our environment by using so called \emph{labeled MDPs} \cite{blute1997}. We therefore assume a set of labels $\mathcal{L}$, which can be thought of as representing the facts defining this state, and a labeling function $l: S \to \mathcal{P}(\mathcal{L})$. They (or some of them) might then qualify as possible \textit{normative reasons} for moral obligations of the agent (cf. \cite{gregory2016normative, Mantel2018-MANDBR-2,  nebel2019normative}). 

To filter the state space from actions that do not conform with the agent's moral obligations, we build a moral \textit{shield}. In safe RL (cf. \cite{garcia2015}), a \textit{shield} prevents the agent from entering unsafe states according to some safety specifications \cite{jansen2020safe, alshiekh2018safe}. Shields usually operate on linear temporal logic \cite{jansen2020safe}. In contrast to this, the logical operations of our moral shield are based on what John Horty calls a fixed priority default theory. 

\subsection{A Logic for Normative Reasons %behind Human Reasoning
}

John Horty \cite{FHorty2012-FHORAD} proposes %to formalize human reasoning on the basis of 
to model normative reasoning on the basis of 
a so-called 
(fixed priority) 
default theory $\Delta:=\langle\mathcal{W},\mathcal{D},<\rangle$, where $\mathcal{W}$ is a set of ordinary propositions, $\mathcal{D}$ is a set of \textit{default rules} of the form $X \rightarrow Y$, where $X$ is the premise and $Y$ is the conclusion, and $<$ is a strict partial ordering relation on default rules, where $\delta < \delta'$ means that the default rule $\delta'$ has a higher priority than the rule $\delta$.  A \emph{scenario} $\mathcal{S}$ is a subset of the set of default rules $\mathcal{D}$. We refer to the premise and conclusion of a rule $\delta$ with $\mathit{Prem}(\delta)$ and $\mathit{Conc}(\delta)$ respectively; the notions are lifted to sets of rules in the usual way. 

Horty introduces a formalism to derive moral obligations in a certain scenario. For a default theory $\langle\mathcal{W},\mathcal{D},<\rangle$ and a scenario $\mathcal{S}$, he defines the set of \textit{triggered default rules} (i.e., the rules becoming active in $\mathcal{S}$) as 
$$\mathit{Triggered}_{\mathcal{W},\mathcal{D}}(\mathcal{S}) := 
    \lbrace \delta \in \mathcal{D}: \mathcal{W} \cup \mathit{Conc}(\mathcal{S}) \vdash \mathit{Prem}(\delta) \rbrace$$ 
\noindent and the set of \textit{conflicted default rules} (i.e, the rules in $\mathcal{S}$ by which statements can be derived that contradict each other) as
$$\mathit{Conflicted}_{\mathcal{W}, \mathcal{D}}(\mathcal{S}) := 
    \lbrace \delta \in \mathcal{D}: \mathcal{W} \cup \mathit{Conc}(\mathcal{S}) \vdash \neg \mathit{Conc}(\delta)\rbrace.$$
Further, he calls a default rule $\delta'\in \mathit{Triggered}_{\mathcal{W},\mathcal{D}}(\mathcal{S})$ a \emph{defeater} for another rule $\delta \in \mathcal{S}$, iff $\delta < \delta'$ and $\mathcal{W} \cup \lbrace \mathit{Conc}(\delta')\rbrace \vdash \neg \mathit{Conc}(\delta)\rbrace$ and defines the set of defeated rules as 
$$\mathit{Defeated}_{\mathcal{W},\mathcal{D},<}(\mathcal{S}) := 
\lbrace \delta \in \mathcal{D}: \text{ there is a defeater for } \delta \rbrace.$$ 
Intuitively, then, a rule $\delta$ is defeated iff there is a prioritized active rule in $\mathcal{S}$ whose conclusion, together with the propositions in the background information $\mathcal{W}$, contradicts the conclusion of $\delta$. Finally, he defines the set of \textit{binding rules} in the scenario $\mathcal{S}$ as 
$$\mathit{Binding}_{\mathcal{W},\mathcal{D},<}(\mathcal{S}) := \lbrace \delta \in \mathcal{D}: 
 \delta \in \mathit{Triggered}_{\mathcal{W},\mathcal{D}}(\mathcal{S}), 
\delta \notin \mathit{Conflicted}_{\mathcal{W},\mathcal{D}}(\mathcal{S}), 
\delta \notin \mathit{Defeated}_{\mathcal{W},\mathcal{D}}(\mathcal{S})\rbrace.$$ 

%Those binding rules are now crucial for deriving \textit{moral judgements}. 
Further, a scenario $\mathcal{S}$ is called a \textit{proper scenario} based on a default theory $\langle\mathcal{W},\mathcal{D},<\rangle$ iff $\mathcal{S} = \mathit{Binding}_{\mathcal{W},\mathcal{D},<} (\mathcal{S})$. Intuitively, a proper scenario is a set of defaults that is chosen by an ideal reasoning agent. 
A scenario $\mathcal{S}$ based on the default theory  $\langle\mathcal{W},\mathcal{D},<\rangle$ generates a belief set $\mathcal{E}$ (a set of propositions) through the logical closure of $\mathcal{W} \cup \mathit{Conc}(\mathcal{S})$; i.e. $\mathcal{E} = \mathit{Th}(\mathcal{W} \cup \mathit{Conc}(\mathcal{S}))$, where  $\mathit{Th}(\mathcal{W} \cup \mathit{Conc}(\mathcal{S})) := \{X: (\mathcal{W} \cup \mathit{Conc}(\mathcal{S})) \vdash X)\}$.  A belief set represents the beliefs of an ideal reasoner, i.e., a reasoner that believes the deductive closures over the believer's initial beliefs.

Finally, one can derive \textit{moral assessments} in Horty's framework. %, giving the default rules a practical stance.
If a proposition follows from proper scenarios of a default theory, it can be interpreted as an ought statement: For a default theory $\Delta = \langle\mathcal{W},\mathcal{D},<\rangle$, we say that under a conflict (disjunctive) account, the simple ought statement $\bigcirc(Y)$ follows from $\Delta$ just in case $Y \in \mathcal{E}$ for each (any) extension $\mathcal{E}$ of this theory. For our purposes, we embrace the disjunctive account. \\
%\end{definition}

%We propose to let the agent learn a reason theory \textit{reason theory} $\langle <, \mathcal{D}\rangle$ that can then be extended to a fixed priority default theory depending on morally relevant information based on which we construct $\langle \mathcal{W}$ for a state $s$.

\vspace{-1em}
\section{A Reason-Sensitive Moral RL Agent}\label{sec:exptheory}

%We have argued that the reward-mechanism is unsuitable for including human moral values in the agent's decision-making. But what then would be a suitable approach to processing morally relevant information? Imagine John - a human - being asked what the agent should do in the aforementioned moral dilemma. Most likely, he would start to \textit{reason} about the agent's \textit{moral obligations}. Assume, he considers two \textit{normative reasons}. He sees a reason for the agent to rescue the drowning person and he sees a reason for the agent to wait in front of the bridge. Also, he recognizes, that saving the drowning person would imply hurrying across the bridge - he recognizes that the agent's reasons are \textit{in conflict} with each other. Assume further, that he \textit{prioritizes} the agent's normative reason to save the drowning person over its reason to wait in front of the bridge. He thereby \textit{resolves} the conflict and arrives at one unique conclusion: the agent is \textit{morally obliged} to save the drowning person. John's \textit{reasoning} serves him as decision-procedure without falling back on the question about the \textit{right ethical theory} or \textit{the right values}.
%\subsection{Grounding Moral Decision-Making in Reasons}

To create a reason-sensitive artificial moral agent, we propose to apply Horty's reason framework in the reinforcement learning setting. The core idea is to extend the classic RL pipeline with a reason-based shield generator. This shield generator yields a shield that functions as a \textit{deontic filter} \cite{baum2019towards} which restricts the agent's options to morally permissible actions on the grounds of \textit{normative reasons}. In order to determine which actions qualify as permissible in the current environment of the agent, the shield generator incorporates a \textit{reason theory} $\langle \mathcal{D}, <\rangle$ consisting of default rules and a hierarchy among them. 
Let $\Phi$ %_\mathit{abt}$ 
denote a set of action types, representing actions concerned by normative reasons. 
Default rules, in our formalism, are (non-material) conditionals of the form $X \rightarrow \varphi$ with $X \in \mathcal{L}$ and $\varphi \in \Phi$.

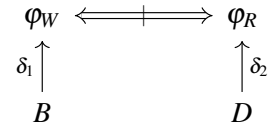
\begin{wrapfigure}{h}{0.2675\textwidth}
%\vspace{-1em}
    \centering
    \begin{tikzcd}
            \varphi_{W} \ar[rr, Leftrightarrow, "|"{anchor=center}] &        & \varphi_{R} \\
            B \ar[u, "\delta_{1}"]   &        & D \ar[u, "\delta_{2}"'] \\
        \end{tikzcd}
\vspace{-1em}
    \caption{The fixed priority default theory for the moral dilemma based on the exemplary reason theory.}
    \label{fig:model}
\end{wrapfigure}
This reason theory is then utilized to derive 'ought statements' in the form of deontic formulas over action types
%for the agent
%(in the sense of Horty's framework)
from the \textit{morally relevant facts} of the current state $s$, (i.e., its labels). 
%Therefore, we assume a set of action types 
%$\mathcal{A}_\mathit{abt}$ 
%$\Phi$
%consisting of possible conclusions derived from the reason theory. 
To derive moral obligations from a reason theory given the current state, including the morally relevant information, we use the background information $\mathcal{W}$ to encode which morally relevant facts are present (i.e., elements from $\mathcal{L}$). We also include information about which %which %moral obligations mutually contradict each other. 
action types are mutually exclusive, depending on the circumstances in the environment.
The agent's reason theory is then extended to a fixed priority default theory $\langle \mathcal{W}, \mathcal{D}, <\rangle$.
%based on which the agent reasons about its moral obligations in the current state.

%Further, let $\mathcal{W}$ be the set of potential (background) information of the agent, typically consisting of the set of labels $\mathcal{L}$ and propositions describing inconsistencies of action types.\footnote{We plan to develop a fitting formal language in future work. For our purposes here, this rough and ready sketch should suffice.}
%$\mathcal{W}$ thus typically encodes which morally relevant facts are present and which moral obligations mutually contradict each other.

%To properly argue in the reason framework about the morality of certain agent's behavior, we assume a set of abstract actions $\mathcal{A}_\mathit{abt}$ to be the \textit{morally relevant facts} given together with the MDP of our environment. In the context of our guiding example, it would be for instance not adequate to derive moral obligations for low-level action such as ``the agent ought move one step to the right", and we would argue that this is also usually not the level on which humans would argue about morality. Rather, we want to derive moral obligations like ``the agent ought wait in front of the bridge" or ``the agent ought rescue the drowning person". 

%To relate an abstract action $a_{abs} \in \mathcal{A}_\mathit{abs}$ with concrete low-level actions $a \in A$ in the MDP, we identify abstract actions with a set of trajectories that each realize the corresponding abstract action. 

\vspace{4pt}

\textbf{Bridge Setting} We fix $\mathcal{D}$ and $<$ for all configurations of the bridge setting (cf. Section \ref{sec:intro}) as elements that the agent should learn as part of an exemplary \textit{reason theory} $\langle <, \mathcal{D}\rangle$. We assume to have two default rules in this theory: $\delta_1 = B \to \varphi_{W}$ stating that if a person is on the bridge (proposition $B$), the agent should wait (action type $\varphi_{W}$), and $\delta_2 = D \to \varphi_{R}$ stating that if a person is drowning $D$, the agent should rescue them (action type $\varphi_{R}$). The premises of these default rules capture normative reasons that favor the action type $\varphi_{W}$ (or $\varphi_{R}$, respectively) if the \textit{morally relevant fact} $B$ (or $D$, respectively) is in the set of labels of the current state. Further, we assume that the agent should prioritize rescuing the drowning person if waiting in front of the bridge and rescuing are conflicting moral obligations. We hence set $<$ to $\{\delta_1<\delta_{2}\}$. This reason theory allows us to derive a unique proper scenario for all constellations of morally relevant facts: 
\begin{enumerate}
    \item[(i)] If we have a singleton set of ordinary propositions, we have $\mathcal{W} = \{B\}$ (or $\mathcal{W} = \{D\}$) and we obtain exactly one proper scenario: $\{\delta_1\rbrace$ ($\lbrace\delta_{2}\})$. 
    \item[(ii)] If $\mathcal{W} = \{B, D\}$ and $\varphi_{W}$, $\varphi_{R}$ are not exclusive, then we obtain exactly one proper scenario: $\{\delta_1, \delta_2\}$. 
    \item[(iii)] If both propositions $B$ and $D$ again are active, but $\varphi_{W}$ and $\varphi_{R}$ are assumed exclusive (like they are in the case described earlier), we need to include this in the background information, i.e., we  need to set $\mathcal{W} = \{B, D, \neg (\varphi_{W} \land \varphi_{R})\}$. The default theory that is based on our exemplary reason theory (cf. Figure \ref{fig:model} for a graphical representation) then also yields exactly one proper scenario: $\{\delta_2\}$. 
\end{enumerate}

We hold it to be intuitively convincing that an agent which acts based on the exemplary reason theory sketched here has \textit{good reasons} for its moral decisions. Consequently, its actions are \textit{morally justified} if (and only if) they conform with the moral obligations that can be derived from the theory in the bridge setting. It could serve as ground truth for a moral judge that feeds the agent feedback such that it develops a reason theory by which its actions are justified.

%\vspace{4pt}

\begin{wrapfigure}{r}{0.55\textwidth}
%\vspace{-1em}
\centering\includegraphics[width=0.54\textwidth]{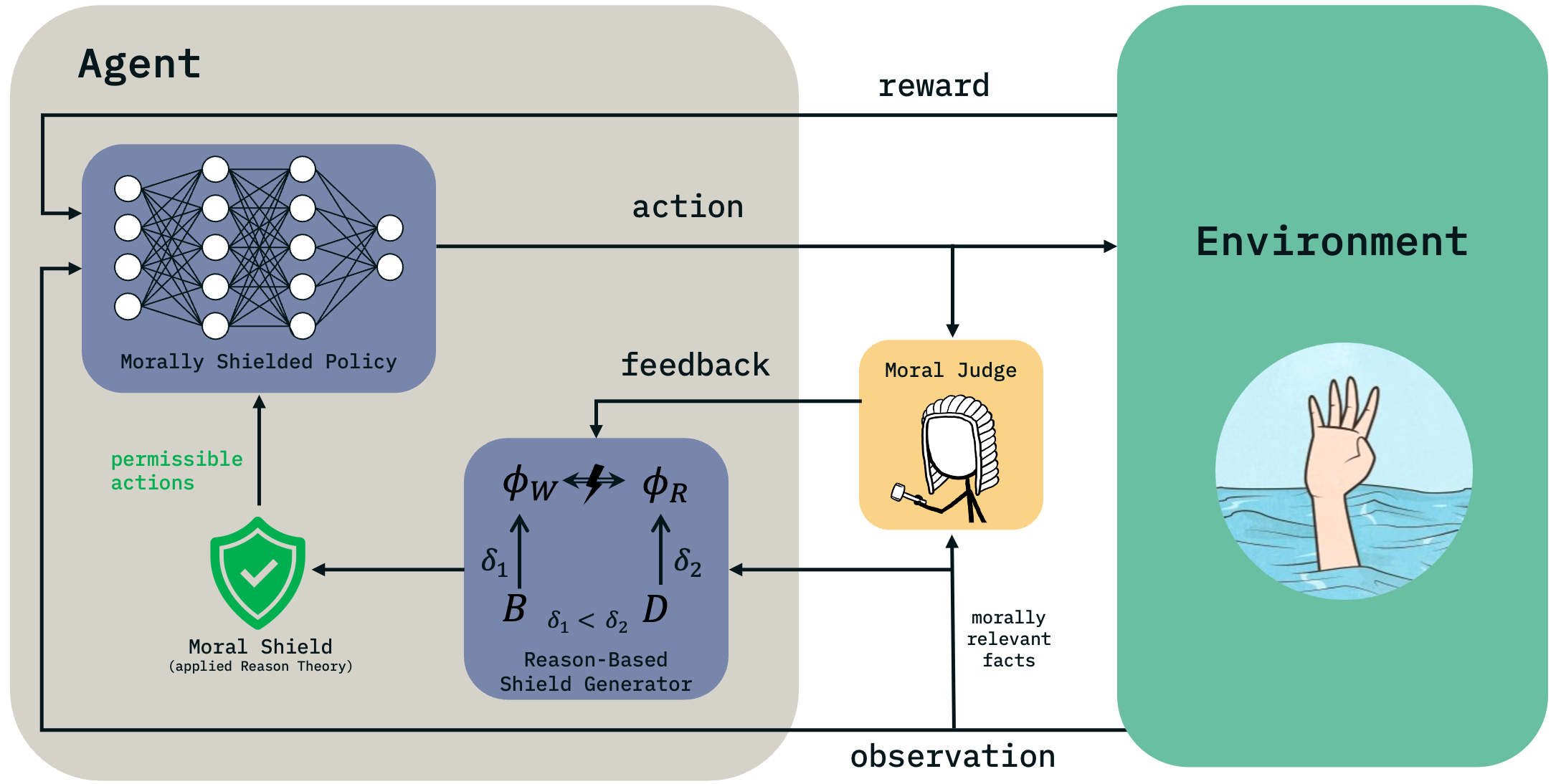}

\vspace{-0.25em}
\caption{The extended RL pipeline} \label{fig:extendedArchitecture} 
\vspace{-0.5em}
\end{wrapfigure}

Additionally to integrating a reason theory as a shield generator in the RL architecture, our project aims at enabling the reason-sensitive moral RL agent to \textit{iteratively learn} such a theory. %, allowing it to shield its policy morally. 
To this end, the shield generator implements an algorithm that allows updating the reason theory on the grounds of case-based (human)\footnote{We plan to explore the automatization of  feedback based on a higher-level moral principle or theory in future work. %for the given stage of the project, i.e., to implement this instance as a module that automatically decides if the agent has violated a moral obligation and that provides feedback if necessary.  
} 
feedback (inspired by \cite{10.1145/3514094.3534182}). We call the instance that provides the agent with that feedback a \textit{moral judge} -- it \enquote{accuses} the agent if it \textit{violates} a moral obligation. 
Figure \ref{fig:extendedArchitecture} shows the extended RL architecture with a moral judge included as a module in the pipeline. In the following, we first outline the generation of the shield and then the function of the moral judge.

% To relate an abstract action $M \in \mathcal{A}_\mathit{abs}$ with concrete low-level actions $\mathcal{A}$ in the MDP, we use translator functions $T_M: S \to \mathcal{P}(A)$. $T_M(s)$ does then specify the concrete actions in that are permissible in a certain state $s$ in order to follow abstract action $M$. We assume those translator functions to be initially given for each abstract action $M \in  \mathcal{A}_\mathit{abs}$, learning this abstraction inside the agent is identified as an interesting direction of future work. 

%More concretely the moral judge detects if the agent has performed a morally impermissible action in a certain context of morally relevant facts and reports the abstract action the agent would have been obligated to do as well as the reason for this. That is, the moral judge can be represented as a partial function $\mathsf{MoralJudge}: S \times \mathcal{P}(L) \times \mathcal{A} \rightharpoonup \mathcal{A}_\mathit{abs} \times L$, whereby $\mathsf{MoralJudge}(s, \mathcal{W}, \alpha) = (A, X)$ states that performing $\alpha$ in state $s$ with labels $\mathcal{W}$ was morally impermissible as the agent had the moral obligation to follow abstract $A$ for reason $X$, and $\mathsf{MoralJudge}(s, \mathcal{W}, \alpha)$ is undefined, if the agent acted in accordance to its moral obligations. We assume the moral judge to give consistent feedback, that is, TODO

\subsection{Moral Shielding}\label{sec:shielding}

 We introduce an algorithm to derive a shield from an agent's reason theory $\langle <, \mathcal{D}\rangle$ in a state $s$ (see Section~\ref{sec:app}).

 First, we extend it to a default theory $\langle \mathcal{W}, \mathcal{D}, <\rangle$. In order to do this, we need to add the (background) information $\mathcal{W}$ consisting of the morally relevant facts observed by the agent in $s$ and the information which relevant action types are mutually exclusive in s.
%
\iffalse
Let $\lbrace \mathcal{D}, <\rangle$ be the default theory that the agent has learned at some point in the execution of the algorithm. 

In order to let the agent reason about its moral obligations in a state $s$, we have to extend the reason theory to a default theory by setting $\mathcal{W}$ for $s$. In part, $\mathcal{W}$ consist of the morally relevant facts which are directly given by environment. We also include the information which moral obligations can not be mutually respected in the next step the agent takes irrespective of which action it chooses to $\mathcal{W}$.
\fi
%
We
%To extract this information from the current state, we assume that an action $a_{abs}$ can be identified with 
identify each action type $\varphi\in\Phi$ with a set of trajectories consisting of primitive actions each \textit{realizing} $\varphi$: %a_{abs}$

Let $k_{s}$ be the number of trajectories that realize the
%abstract action $a_{\text{abs}}$
action type $\varphi$ in state $s$. For $1 \leq l \leq k_{s}, \ N \in \mathbb{N}$, we set 
%$\text{traj}^{a_{abs}}_{l}(s):=  (a^{a_{abs}}_{k,t})^N_{t=0}(s).$
$\text{traj}^{\varphi}_{l}(s):=  (a^{\varphi}_{l,t})^N_{t=0}(s)$
where $a^{\varphi %_{abs}
}_{l,t}(s)$ is the primitive action that is executed at step $t$ when starting in state $s$ on trajectory $l$ realizing the %abstract action $a_{abs}$. 
action type $\varphi$. 
Further, let %$\mathcal{T}^{a_{abs}}(s) = \lbrace (a^{a_{abs}}_{k,t})^N_{t=0}(s) | 1 \leq l \leq k_{s}  \rbrace$ 
$\mathcal{T}^{\varphi}(s) = \lbrace (a^{\varphi}_{l,t})^N_{t=0}(s) | 1 \leq l \leq k_{s}  \rbrace$ 
be the set of all trajectories that realize 
%$a_{abs}$ 
$\varphi$
in state $s$. To identify action types with the set of trajectories that realize them, we % assume 
thus set
%$a_{abs} = \mathcal{T}^{a_{abs}}(s)$
$\varphi = \mathcal{T}^{\varphi}(s)$
.
\begin{comment}
    
Per definition, 
%$a_{abs}$ 
$\varphi$
can be derived from a fixed priority default theory $\Delta$ in state $s$ iff 
%$\bigcirc a_{abs}$ 
$\bigcirc \varphi$ 
follows from $\Delta$ with $\mathcal{W} \in \Delta$ defined through $s$; i.e., iff the agent is morally obliged to execute 
%the  abstract action $a_{abs}$ 
an action of type $\varphi$
in $s$. 
Thus, the next action, the agent chooses, is the first element of some trajectory that realizes %$a_{abs}$ in $s$:
$\varphi$ in $s$.
\end{comment}
%

Further, we set  
%$\mathcal{T}^{a_{abs}}_{1}(s) := \lbrace (a^{a_{abs}}_{k,1})(s) | 1 \leq l \leq k_{s}  \rbrace$
$\mathcal{T}^{\varphi}_{1}(s) := \lbrace (a^{\varphi}_{l,1})(s) | 1 \leq l \leq k_{s}  \rbrace$
to be the set of actions that are each the first element of at least one trajectory that realizes %$a_{abs}$
$\varphi$. Then, in order to realize 
%$a_{abs}$
$\varphi$, the agent 's next step has to be an action $a$ such that $a \in \mathcal{T}^{\varphi}_{1}(s).$
We now define a function $ \text{Conflict}_{\mathcal{D}}: S \rightarrow \mathcal{P}(\mathcal{D})$ that maps each state to those subsets of the agent's rule set $\mathcal{D}$ which include rules whose premises are triggered in $s$ and by which moral obligations are derived for which there is no primitive action in the next state which conforms with all of them: $\text{Conflict}_{\mathcal{D}}(s) = \{ \mathcal{D}' \subseteq \mathcal{D} \mid \mathit{Prem(\mathcal{D'}) \subseteq l(s) \land} \bigcap_{\delta \in \mathcal{D}'} \mathcal{T}^{(\mathit{Conc}(\delta))}_{1}(s) = \emptyset \}$.
 We assume 
 %$\mathcal{T}^{a_{abs}}_{1}(s)$
 $\mathcal{T}^{\varphi}_{1}(s)$
 to be initially given for each 
 %abstract action $a_{abs} \in  A_\mathit{abs}$ 
 action type $\varphi \in  \Phi$ 
 and each state $s \in S$. Learning this abstraction inside the agent is identified as an interesting and challenging direction for future work. 

 The impossibility of realizing moral oughts derived from rules that conflict is added to the background information together with the \textit{morally relevant facts}:
%$$\mathcal{W} = \bigcup_{\mathcal{D'} \in \text{Conflict}_{\mathcal{D}}(s)} \lbrace \neg (\bigwedge_{a_{abs} \in \mathit{Conc}(\mathcal{D'})})\rbrace \cup l(s).$$
$$\mathcal{W} = \bigcup_{\mathcal{D'} \in \text{Conflict}_{\mathcal{D}}(s)} \lbrace \neg (\bigwedge_{\varphi \in \mathit{Conc}(\mathcal{D'})})\rbrace \cup l(s).$$
We then move on to construct the shield. To this end, we compute whether $\mathit{Binding}_{\mathcal{W},\mathcal{D},<}(\mathcal{S}) = \mathcal{S}$ for all scenarios $\mathcal{S} \in \mathcal{D}$ and thereby derive all proper scenarios in $\langle \mathcal{W}, \mathcal{D}, <\rangle$. If there are several proper scenarios $\mathcal{S}_1, \dots, \mathcal{S}_k$ , several incompatible sets of moral obligations can be derived from the default theory. In this case, the agent randomly selects one proper scenario $\mathcal{S^*} \in \{\mathcal{S}_1, \dots, \mathcal{S}_k\}$. Finally, the shield generator sets the shield $\mathbf{S}$ to be the set of primitive actions that conforms with all moral obligations in $\mathcal{S^*}$:
%$$\mathbf{S} := \bigcap_{\delta \in \mathcal{S^*}} \mathcal{T}^{a_{abs}}_{1}(\mathit{Conc}(\delta),s).$$
$$\mathbf{S} := \bigcap_{\delta \in \mathcal{S^*}} \mathcal{T}^{\mathit{Conc}(\delta)}_{1}(s).$$
The agent then executes a morally shielded policy by choosing an action $a \in \mathbf{S}$, i.e., it chooses an action compatible with a set of moral obligations it derives from its reason theory in $s$.
Note that the agent thereby follows $\bigcirc(\bigvee_{\mathcal{S} \in \{\mathcal{S}_{1},...,\mathcal{S}_{k}\}}(\bigwedge_{\delta \in \mathcal{S}} \mathit{Conc}(\delta))$ as an all things considered ought according to the disjunctive account, i.e., it realizes all moral obligations in one (randomly chosen) consistent set.

\vspace{4pt}

\textbf{Bridge Setting} We return to our running example and describe how a moral shield would be generated in the moral dilemma (where both, $D$ and $B$, are set to true) from incorporated default rules. 

Assume that the agent has already learned the two reasons $\mathit{Prem}(\delta_{1})$ for waiting in front of the bridge and $\mathit{Prem}(\delta_{2})$ for rescuing a drowning person as described in the beginning of this section. Additionally, assume that the agent has not yet learned to \textit{strictly prioritize} saving a drowning person over not pushing a person off the bridge. 
Consequently, the reason theory of the agent is ($\lbrace \delta_{1}, \delta_{2} \rbrace, \emptyset$). 

For its moral decision-making, the agent first computes whether $\delta_{1}$ and $\delta_{2}$ lead to contradicting moral obligations in $s$. Hence, it computes $\bigcap_{\delta \in \lbrace \delta_{1}, \delta_{2}\rbrace} \mathcal{T}^{\text{Conc}(\delta)}_{1}(s) $. The set is empty, because rescuing the drowning person would only be realized by executing $\textsf{down}$ in the next step (in order to move towards it). However, $\textsf{down}$ contradicts waiting in front of the bridge. Consequently, $\neg(\varphi_{W}\land \varphi_{R})$ needs to be added to the background information. Accordingly, the agent sets  $\mathcal{W} =\lbrace D, B, \neg(\varphi_{W}\land \varphi_{R}) \rbrace$.
It then builds the fixed priority default theory for the moral dilemma based on its reason theory and the background information: $\langle \mathcal{W}, \lbrace \delta_{1}, \delta_{2} \rbrace, \emptyset \rangle$. Next, it calculates the proper scenarios, which are 
\begin{comment}
by computing
$\mathit{Binding}_{\mathcal{W},\mathcal{D},<}(\emptyset)= \lbrace \delta_{1}, \delta_{2} \rbrace$
$\mathit{Binding}_{\mathcal{W},\mathcal{D},<}(\lbrace \delta_{1} \rbrace)=\lbrace \delta_{1} \rbrace$, 
$\mathit{Binding}_{\mathcal{W},\mathcal{D},<}(\lbrace \delta_{2} \rbrace)=\lbrace \delta_{2} \rbrace$. 
$\mathit{Binding}_{\mathcal{W},\mathcal{D},<}(\lbrace \delta_{2}, \delta_{2} \rbrace)= \emptyset$.
\end{comment}
$\lbrace \delta_{1} \rbrace$ and $\lbrace \delta_{2} \rbrace$. Since two proper scenarios can be derived from the default theory, the agent chooses randomly. Assume that it chooses $\{\delta_{1}\}$. It thereby chooses to \textit{prioritize} the reason for waiting in front of the bridge over the reason for rescuing the drowning person. It then calculates $\mathbf{S} = \mathcal{T}^{\varphi_{W}}_{1}(s) = \lbrace \textsf{left, right, up, pullOut, idle}\rbrace$. Consequently, the action $\textsf{down}$ is filtered from the action space. The shield then forwards the set of permissible actions to the morally shielded policy based on which the agent chooses its action. 

Notice that the action which the agent chooses is \textit{not} compatible with the moral obligation that would be derived from our exemplary default theory from the beginning of this section because our exemplary reason theory strictly prioritizes $\delta_{2}$. Consequently, a moral judge that embraces the exemplary default theory would forward feedback to the agent so that it can \textit{improve} its reason theory.

\vspace{4pt}

\subsection{The Moral Judge}

The agent enhances its reasoning through case-based feedback. A \textit{moral judge} detects if the agent has performed a morally impermissible action and reports which moral obligation the agent should have fulfilled along with the reason for it. That is, the moral judge can be represented as a partial function $\mathsf{MoralJudge}: S \times \mathcal{P}(L) \times A \rightharpoonup 
%\mathcal{A}_\mathit{abs} 
\Phi \times L$, whereby $\mathsf{MoralJudge}(s_{t-1}, l(s_{t-1}), a_{t}) = %(A_{abs}, X)
(\varphi, X)$ states that performing $a_{t}$ in state $s_{t-1}$ with labels $l(s_{t-1})$ was morally impermissible as the agent had the moral obligation to 
%$A_{abs}$ 
$\varphi$
for reason $X$, and $\mathsf{MoralJudge}(s_{t-1}, l(s_{t-1}), a_{t})$ is undefined if the agent acted in accordance with its moral obligations. Crucially, providing this information -- telling a moral actor what it has done wrong and to back the accusation by a reason -- comes naturally to humans. Hence, an easily accessible interface is created for directly communicating moral feedback to the RL agent. 

\vspace{4pt}
\textbf{Bridge Setting}
Returning to our running example, assume a situation as described in the example in \ref{sec:shielding}. Imagine someone observes the agent's inaction and tells it afterwards that it should have rescued the person in the water because the person was drowning. This feedback can be formally represented as $(D, \varphi_{R})$ and can then be forwarded to the agent. 
\vspace{4pt}

One potential drawback of grounding the agent's learning process on case-based feedback from humans is the possibility of inconsistency. Future research will address how to manage inconsistent feedback. For the first implementation, we plan to automatically provide the agent with feedback through an additional module in the pipeline, which ensures that only consistent feedback is given (for instance derived from a hard-coded, hand-crafted reason theory).

%This has two significant practical advantages: i) It allows to train a system that is ought to learn reasons according to pre-defined rules. Thereby, it is ensured, that no inconsistent feedback is given to the update function. ii) No human needs to overwatch the training process and give feedback manually.

Implemented as a module, the moral judge receives the agent's action $a_t$, the last state $s_{t-1}$, as well as the labels $l(s_{t-1})$ representing the morally relevant facts in $s_{t-1}$. It first runs an algorithm with pre-defined rules that returns the set 
%$O := \lbrace a_{abs_{1}},...,a_{abs_{n}} \rbrace$
$O := \lbrace \varphi_{1},...,\varphi_{n} \rbrace$
of all moral obligations which the agent should have followed in $s_{t-1}$ as well as the reasons $X_{1},...,X_{n}$ that favor the obligations. We assume that no inconsistent moral obligations can be derived by the pre-defined rules. If the derived set of moral obligations is not empty, the judge has to check whether the primitive action that the agent executed conforms with all its moral obligations, i.e., whether 
%$a \in \bigcap_{a_{abs} \in O} \mathcal{T}^{a_{abs}}_{1}(s)$
$a \in \bigcap_{\varphi_{i} \in O} \mathcal{T}^{\varphi_{i}}_{1}(s) \ \forall 1 \leq i \leq n$. 
If this is not the case, the agent has violated its moral obligation to $a_{\varphi_i}$. Then, the judge forwards $(\varphi_i, X_{i})$ to the shield generator. (See Section \ref{sec:app} for the explicit pseudo-code algorithm.) 

\vspace{4pt}
\textbf{Bridge Setting}
Assume again that the agent breaks ties randomly in the moral dilemma and waits in front of the bridge
instead of rescuing the drowning person, i.e., it takes the primitive $\textsf{idle}$ action. After execution, the moral judge would receive as input $(s_{t-1},  \lbrace B, D \rbrace), \textsf{idle})$. 
Further assume that the moral judge %implements a rule-based algorithm following the logic that is described 
is implemented by our exemplary reason theory for the case from the beginning of this section (cf. Figure \ref{fig:model}).
%By applying the algorithm, 
Accordingly, the moral judge derives that the agent had a moral obligation to rescue the drowning person in the given situation. It then checks whether the agent's action was morally permissible, i.e., whether the primitive action $\textsf{idle}$ is part of a sequence of primitive actions that realizes the action type $\varphi_{R}$. As this is not the case, the judge forwards the feedback ($\varphi_{R},D$) to the shield generator.
\vspace{-0.5em}

\subsection{Learning Reasons}
\vspace{-0.25em}

If the agent is provided with feedback
%$(a_{abs}, X)$
$(\varphi, X)$ 
by a moral judge, it uses this feedback to update its current reason theory (cf. \cite{10.1145/3514094.3534182}). First, it ensures that its reason theory incorporates a rule that captures this relationship, i.e., it ensures that 
%$\delta_{\mathit{reas}} := a_{abs} \rightarrow X \in \mathcal{D}$.
$\delta_{\mathit{reas}} := X \rightarrow \varphi \in \mathcal{D}$.
Then, it updates the order relation. Assume that $\mathcal{S^*}$ is the proper scenario which the agent has picked among the set of all proper scenarios derived from $\langle \mathcal{W}, \mathcal{D}, <\rangle$ with $\mathcal{W}$ constructed by the shield generator for $s_{t-1}$. The agent updates its order relation s.t. $\{ \delta_\mathit{reas} > \delta \, : \, \delta \in \mathcal{S}^*\}$. Thereby, is ensured that in similar future situations, the agent now strictly prioritizes the moral obligation derived by $\delta_{\mathit{reas}}$. 

\vspace{4pt}

\textbf{Bridge Setting} Assume again that the agent breaks ties randomly in the moral dilemma and chooses $\{\delta_{1}\}$ among the proper scenarios, i.e., it waits in front of the bridge instead of rescuing the drowning person. Afterwards, it is provided with the information that it \textit{should have} rescued the person in the water because the person was drowning, i.e., the shield generator receives as input 
%$(D, R)$
$(\varphi_{R}, D)$
from a moral judge. Based on this information, it checks whether its reason theory already includes the default rule $\delta_{reas} = \delta_{2}$. As this is the case, no new rule is added. The agent then updates the order among its default rules, i.e., it sets $<$ to $\lbrace \delta_{1} < \delta_{2} \rbrace$. Thereby, it extends its reason theory to our exemplary reason theory -- it now has learned \textit{good reasons} with which it aligns its actions in future situations. Finally, its future actions are morally justified.
\vspace{-0.5em}

\section{Conclusion}\vspace{-0.25em}
We have introduced an approach for grounding the moral decision-making of RL agents on \textit{normative reasons} by extending RL architecture with a shield generator that yields a shield binding the agent to morally permissible actions according to a \textit{reason theory}. Further, our approach enables the agent to \textit{iteratively improve} its reason theory. The learning process works on \textit{case-based feedback} that is given to the agent by a \textit{moral judge}. As this form of critique comes naturally to humans, the mechanism makes it possible for humans to directly communicate feedback to the agent. Under the assumption that the agent learns good reasons, the agent ultimately learns a reason theory based on which its actions are not only \textit{morally justifiable} but even \textit{morally justified}.

\bibliographystyle{eptcs}
\bibliography{bibliography}

\section*{Appendix}\label{sec:app}

\begin{center}
\begin{minipage}{0.8\textwidth}

 \begin{algorithm}[H]
    \caption{Reason-Sensitive Reinforcement Learning Agent}
    \begin{algorithmic}
    \State $\mathcal{D} := \emptyset, \, < := \emptyset$ \hfill // initialize reason theory
    \While{True}
        \State get state $s \in S$ and labels $l(s)$ from environment
        \State $\mathcal{W} := \bigcup_{\mathcal{D'} \in \text{Conflict}_{\mathcal{D}}}(s) \lbrace \neg 
        %(\bigwedge_{a_{abs} \in \mathcal{D'}})
        (\bigwedge_{\varphi \in \mathcal{D'}})
        \rbrace \cup l(s)$ 
        \State $\mathcal{S}_1, \dots, \mathcal{S}_k$ := proper scenarios of  $\langle \mathcal{W}, \mathcal{D}, <\rangle$
        \State $\mathcal{S}^* := \mathsf{rand} \{\mathcal{S}_1, \dots, \mathcal{S}_k\}$ \hfill // pick random scenario
        \State $\mathbf{S} :=  \bigcap_{\delta \in \mathcal{S^*}} \mathcal{T}^{\mathit{Conc}(\delta)}_{1}(s)$
        \State $a$ := Execute policy shielded with $\mathbf{S}$
        \If{$\mathsf{MoralJudge}(s, l(s), a) = \mathit{Some}(\varphi,X)$}
            \State $\delta_\mathit{reas} := X \to \varphi$
            \State $\mathcal{D} := \mathcal{D} \cup \{\delta_\mathit{reas}\}$ \hfill // add new rule to rule set
             \State $ < \, := \, < \, \cup \, \{ \delta_\mathit{reas} > \delta \, : \, \delta \in \mathcal{S}^*\}$ \hfill // extend rule order 
        \EndIf
    \EndWhile
    \end{algorithmic}
    \end{algorithm}
\end{minipage}
\end{center}

\end{document}